\pdfoutput=1
\documentclass[11pt]{article}

\usepackage[]{ACL2023}

\usepackage{times}
\usepackage{latexsym}

\usepackage{float}
\restylefloat{table}
\usepackage{placeins}
\usepackage{graphicx}
\usepackage{hyperref}
\usepackage{comment}

\usepackage[T1]{fontenc}
\usepackage[utf8]{inputenc}

\usepackage{microtype}

\usepackage{inconsolata}

\title{A Pipeline to Assess Merging Methods via Behavior and Internals}

\author{Yutaro Sigrist\thanks{*Corresponding author \href{yutarosigrist@gmail.com}{yutarosigrist@gmail.com}} $^{1}$ and
Andreas Waldis$^{1,2}$\\
 $^1$Information Systems Research Lab, Lucerne University of Applied Sciences and Arts \\
$^2$Ubiquitous Knowledge Processing Lab (UKP Lab), 
Technical University of Darmstadt\\
\texttt{\href{http://www.hslu.ch/}{www.hslu.ch}} \hspace{0.5em} \texttt{\href{http://www.ukp.tu-darmstadt.de/}{www.ukp.tu-darmstadt.de}} \\
}

\begin{document}
\maketitle
\begin{abstract}

Merging methods combine the weights of multiple language models (LMs) to leverage their capacities, such as for domain adaptation. 
While existing studies investigate merged models from a solely behavioral perspective, we offer the first comprehensive view by assessing and connecting their behavior and internals. 
We present a novel evaluation pipeline that first merges multiple parent LMs and then evaluates the merged models in comparison to the initial ones based on their behavior on downstream tasks, like MMLU, and the internal encoded linguistic competence.
We showcase this pipeline by assessing the merging of instruction fine-tuned with math- and code-adapted LMs from the \texttt{Qwen2.5} family. 
Our results show that merging methods impacts behavior and internals differently. 
While the performance of merged models is typically between that of the two parent models, their encoded information about linguistic phenomena – particularly in \textit{morphology} and \textit{syntax} – can surpass the parent models.
Moreover, we find weak ranking correlation between this behavior and internal evaluation. 
With our pipeline and initial results, we emphasize the need for more comprehensive evaluations of model merging methods to gain a faithful understanding of their capabilities and reliability, beyond potential superficial behavioral advances.

\end{abstract}

\section{Introduction}

With the rise of competitive open-weight models \citep{Dubey2024TheL3,jiang2024mixtral}, adapting large language models (LLMs) to specific use cases has become common practice. One approach, fine-tuning, enables adaptation to particular domains, such as mathematics or coding \cite{lewkowycz2022solvingquantitativereasoningproblems}.
Still, it suffers from high computational costs and the risk of catastrophic forgetting \citep{DBLP:journals/corr/abs-2308-08747}. When a model must handle a variety of tasks, fine-tuning becomes more complex \cite{lu2024finetuninglargelanguagemodels}. Model merging offers methods that overcome some of these issues. These methods combine multiple models into a single one, thereby enhancing performance across the tasks for which the individual parent models were fine-tuned \citep{yang2024modelmergingllmsmllms}.

Existing work in model merging primarily focuses on techniques to improve the performance of merged models. For example, \citet{lu2024finetuninglargelanguagemodels} explored strategies for merging domain-adapted parent models, such as those in materials science and engineering, to efficiently transfer this domain adaptation to other models. 
In addition, \citet{dziadzio2024mergemultimodalmodelstime} introduces the concept of temporal model merging, addressing the challenge of integrating knowledge from multiple parent models trained on various tasks over time. Furthermore, \citet{goddard2025arceesmergekittoolkitmerging} introduces MergeKit, a merging toolkit supporting various methods.

\begin{figure*}[ht]
    \centering
    \includegraphics[width=1\linewidth]{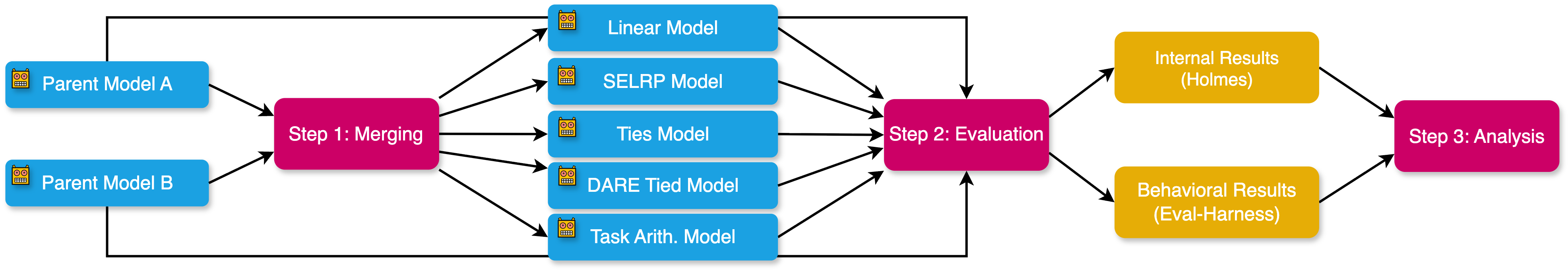}
    \caption{This flowchart shows a model evaluation pipeline. Input "Instruct" and "Math/Coder" models undergo a "Merging" process (using techniques like \texttt{Linear, SLERP, TIES, DARE TIES, Task Arithmetic}). The merged output is then evaluated by "Holmes-Evaluation (Flash-Holmes)" for linguistic competencies (reasoning, morphology, etc.) and "Evaluation LM-Harness" using benchmarks (\texttt{Leaderboard}). The pipeline concludes with analyzing the correlation between Harness and Holmes.}
    \label{fig:piepline}
\end{figure*}

While this variety of studies highlights the popularity of model merging, existing work focuses primarily on assessing these models based on their behavior. As a result, we lack a comprehensive evaluation that assesses and connects both model behavior and internals. This research gap means we risk relying on potentially unstable merged models based on their seemingly better outputs. With this work, we address this shortage in model merging research and ask the following question:
\begin{quote}
\textit{How does model merging affect the internal representations of language models?}
\end{quote}
To answer this, we present an evaluation pipeline (\autoref{sec:pipeline}) with three stages: \textbf{1)} we merge multiple parent models with different strategies, \textbf{2)} we evaluate both the behavior and internals of the parent and resulting merged models, and \textbf{3)} we connect the findings from both evaluations. We showcase this pipeline through experiments on the \textit{Qwen2.5} model family \citep{qwen2025qwen25technicalreport}, which merges instruction fine-tuned and domain-adapted models. Our results show that while merged models generally perform between the parent models, their encoded linguistic competence—particularly for \textit{morphology} and \textit{syntax}—can be stronger than the parent ones'. At the same time, weak rank correlations between behavior and internals underscore the need for comprehensive evaluations, as single perspectives are insufficient to capture the full complexity of language models \citep{hu-levy-2023-prompting,waldis2025aligned}, for and beyond model merging.

With this work, we provide the ground for more comprehensively assessing model merging methods and resulting models by contributing as follows:
\begin{itemize}
\item We introduce the first pipeline that combines model merging with a comprehensive evaluation of model behavior and internals.
\item We present initial insights into connecting the behavioral and internal interpretability perspectives in the context of domain adaptation scenarios.
\item We make all code, including pipeline and analysis, online available.\footnote{\href{https://github.com/yusigrist/LLM-Merging-Piepline}{https://github.com/yusigrist/LLM-Merging-Piepline}}
\end{itemize}

\section{Related works}
\paragraph{Model Merging}

Model merging enables the efficient combination of specialized models \citep{yang2024modelmergingllmsmllms}.
This includes simpler methods like linearly averaging multiple parent models \cite{wortsman2022modelsoupsaveragingweights} or spherical interpolation (SLERP)\footnote{\href{https://github.com/Digitous/LLM-SLERP-Merge}{https://github.com/Digitous/LLM-SLERP-Merge}}.
More sophisticated methods try to locate task-specific regions to better preserve specific skills of the parent models, popular examples are \textit{Task Arithmetic} \citep{ilharco2023editingmodelstaskarithmetic}, \textit{Ties} \citep{yadav2023tiesmergingresolvinginterferencemerging}, or \textit{Dare-Ties} \citep{yu2024languagemodelssupermario}.
The general popularity of such methods is evident in the development of toolboxes that easily merge two or more parent models, such as \textit{MergeKit} \citep{goddard-etal-2024-arcees}, resulting in thousands of merged models available on Huggingface.

\paragraph{Model Evaluation}
A predominant line of research focuses on evaluating language models based on their behavior. 
This includes general language understanding (GLUE; \citealt{wang-etal-2018-glue} or SuperGLUE; \citealt{10.5555/3454287.3454581}), question answering (SQuAD; \citealt{rajpurkar-etal-2016-squad}), or more broad evaluations, as done in HELM \citep{liang2022holistic} or \textit{evaluation harness} \citep{eval-harness}.
Moreover, research also focused on comprehensively assessing specific domains, like factuality \citep{chen2023felm,DBLP:conf/eacl/MuhlgayRMLRBALSS24}, medical texts \citep{bedi2025medhelm}, or legal reasoning \citep{guha2023legalbench}.
While these benchmarks focus on model behavior, there has been little comprehensive work addressing model internals.
\citet{conneau-etal-2018-cram} introduce a benchmark with ten tasks to assess linguistic properties of sentence representations, \citet{warstadt-etal-2020-blimp-benchmark} presents a collection of minimal pairs to examine whether language models can internally differentiate between linguistic acceptable and unacceptable sentences, and \citet{waldis2024holmesbenchmarkassesslinguistic} introduced, with \texttt{Holmes}, a benchmark that studies the linguistic competence of language models based on their internals across five linguistic phenomena and more than 160 distinct probing tasks. 

With this work, we extend the evaluation scope of merging methods to model internals, a previously understudied aspect in model merging methods, in favor of a more comprehensive understanding of how these approaches combine model weights, and offer the base to assess how this changes information flow within models. 

\begin{table*}[t]
\centering
\begin{tabular}{lllllll}
\hline
\texttt{Feature} &
  \texttt{Linear} &
  \texttt{SLERP} &
  \texttt{Task Arithmetic} &
  \texttt{TIES} &
  \texttt{DARE TIES}\\ \hline
\texttt{Complexity} & 
  low & 
  moderate & 
  moderate & 
  moderate & 
  high \\
\texttt{\begin{tabular}[c]{@{}l@{}}Power usage\\ (Inference)\end{tabular}} &
  low &
  low &
  low &
  low &
  low \\
\texttt{\begin{tabular}[c]{@{}l@{}}Power usage\\ (Merging)\end{tabular}} & 
  low & 
  low & 
  low & 
  low & 
  low \\
\texttt{Multi-Model} & 
  yes & 
  2 models & 
  yes & 
  yes & 
  yes \\
\texttt{Performance} &
  simple &
  \textgreater{}Linear &
  \textgreater{}Linear &
  \textgreater{}SLERP &
  \textgreater{}SLERP \\
\texttt{\begin{tabular}[c]{@{}l@{}}Mergekit \\ Difficulty \end{tabular}} & 
  very low & 
  low & 
  low & 
  moderate & 
  high \\ \hline
\end{tabular}
\caption{\label{citation-guide}
Comparison of a set of Features for five different merging methods (\texttt{Linear, SLERP, Task Arithmetic, TIES, DARE TIES}). 
The Performance always depends on the configuration and model. So in some cases \texttt{Linear} can be better than \texttt{DARE TIES}.
}
\end{table*}

\section{The Pipeline}\label{sec:pipeline}
With the presented pipeline (Figure \ref{fig:piepline}), we offer a flexible workflow to merge parent models, and evaluate these parents as well as the resulting merged models from a behavioral and internal perspective. 
Finally, we connected these distinct interpretability perspectives to gain a more in-depth understanding of model workings. 

\subsection{Merging methods}
In a first step, this pipeline uses \textit{MergeKit} \citep{goddard-etal-2024-arcees} to combine two parent models. 
Table \ref{citation-guide} provides an overview of these methods, which vary in complexity, and compares them based on a few essential features. Note that performance always depends on the configuration and model, such that \texttt{Linear} can work better than \texttt{DARE TIES} under certain conditions.

\paragraph{Linear} averages corresponding weights of multiple models, often fine-tuned from a common base with different hyperparameters. This method stands out because of its low complexity and very low difficulty when implemented with Mergekit, making it highly accessible to practitioners. The approach maintains low power usage during both inference and merging processes, while supporting multi-model combinations effectively. Key Application/Strength are simplicity, improved accuracy, and robustness without an increase in inference costs for similar models. Despite its simplicity, performance can vary significantly depending on configuration and, in some cases, may be limited compared to more complex methods \cite{wortsman2022modelsoupsaveragingweights}.

\paragraph{SLERP}
interpolates weights between two models along spherical paths, with the aim of a smoother blend than \texttt{Linear}. This method operates with moderate complexity while maintaining low power consumption during both inference and merging. However, it is limited to only two models in practice, making it less suitable for multi-model scenarios. The implementation difficulty of Mergekit remains low, making it accessible for most applications. Key Application/Strength are effective two-model merging, potentially finding lower loss barrier paths. Performance typically exceeds linear methods when properly configured, offering a balanced approach between simplicity and effectiveness \cite{Animating_rotation_with_quaternion_curves, NVIDIATechnicalBlog_2024}.

\paragraph{Task Arithmetic}
computes "task vectors" (fine-tuned base weights) and adds them to a base model to combine capabilities. This approach maintains moderate complexity while effectively supporting multi-model merging scenarios. The method benefits from low power usage during both inference and merging phases, with relatively low implementation difficulty in Mergekit frameworks. Key Application/Strength are flexible multitasking combination in model editing. Performance generally surpasses \texttt{Linear} approaches, providing a practical solution to combine diverse model capabilities without significant computational overhead \cite{ilharco2023editingmodelstaskarithmetic}.

\paragraph{TIES}
trims insignificant parameter changes, selects a dominant sign for conflicts, and merges only aligned parameters. This method introduces high complexity in its implementation, which requires careful configuration and understanding of the interaction of the parameters. Despite the increased complexity, it maintains low power usage during inference and merging, while supporting multi-model combinations. Key Application/Strength are the reduction of parameter interference, especially for sign conflicts, in multi-model merging. Performance typically exceeds \texttt{SLERP} methods when properly implemented, making the additional complexity worthwhile for demanding applications \cite{yadav2023tiesmergingresolvinginterferencemerging}.

\paragraph{DARE TIES}
uses the DARE (Drop And REscale) scheme to sparsify task vectors (randomly drops and rescales parameters) before applying \texttt{TIES}-merging. This advanced method operates with high complexity, building on the TIES framework with additional parameter sparsification techniques that require expertise in both \texttt{DARE and TIES} methodologies. Power consumption remains low during both inference and merging, while supporting comprehensive multi-model integration. Key Application/Strength are the further mitigated interference by reducing parameter density before \texttt{TIES}.  Performance generally surpasses \texttt{SLERP} methods and can exceed \texttt{TIES} in many configurations, representing a state-of-the-art approach to model merging \cite{yadav2023tiesmergingresolvinginterferencemerging, yu2024languagemodelssupermario}.

\subsection{Evaluation}
In a second step, our pipeline automatically evaluates both the parent models and the resulting merged ones.
This includes a behavioral evaluation using the eval-harness library \citep{eval-harness} and the streamlined and efficient version of \texttt{Holmes} \citep{waldis2024holmesbenchmarkassesslinguistic}, which comprehensively assesses information about linguistic phenomena within model internals. 

\subsubsection{LM-Harness-Evaluation}

The LM-Harness-Evaluation is a comprehensive evaluation framework to assess language models \citep{eval-harness}. 
It includes a variety of tasks in favor of a standardized and reproducible evaluation of these models. For our study, we use the following tasks also used in the OpenLLM evaluation leaderboard \citep{open-llm-leaderboard-v2}:
\begin{itemize}
        \item \texttt{BBH}: Collection of LLM tasks across domains, for example, language understanding, mathematical reasoning, common sense, and world knowledge.
        \item \texttt{math hard}: High school level competitions for mathematical problems: complex algebra, geometry problems, etc.
        \item \texttt{MUSR}: Reasoning on and understanding of long texts: language understanding, reasoning capabilities
        \item \texttt{GPQA}: PhD-level knowledge multiple choice questions in science: Chemistry, Biology, and Physics
        \item \texttt{MMLU-PRO}: Expert reviewed multiple choice questions across domains, for example: medicine and healthcare, law and ethics, engineering, mathematics
\end{itemize}

\subsubsection{Holmes-Evaluation}

Using \texttt{Holmes}, we evaluate the internal representations using classifier based probing \citep{belinkov-2022-probing}
Specifically, we use \texttt{Flash-Holmes}, a streamlined and efficient version of the benchmark that preserves the effectiveness of the evaluation, even substantially reducing the number of instances.
It assesses a range of linguistic phenomena, including \texttt{morphology, syntax, semantics, reasoning, and discourse}, using more than 160 unique probing tasks. This involves training simple linear classifiers on the internal representations of the models' last layer to predict specific linguistic properties, thereby assessing how well the model encodes this information
\cite{hewitt2019designinginterpretingprobescontrol, voita2020informationtheoreticprobingminimumdescription, waldis2024holmesbenchmarkassesslinguistic}.

\begin{figure*}
    \centering
    \includegraphics[width=1\linewidth]{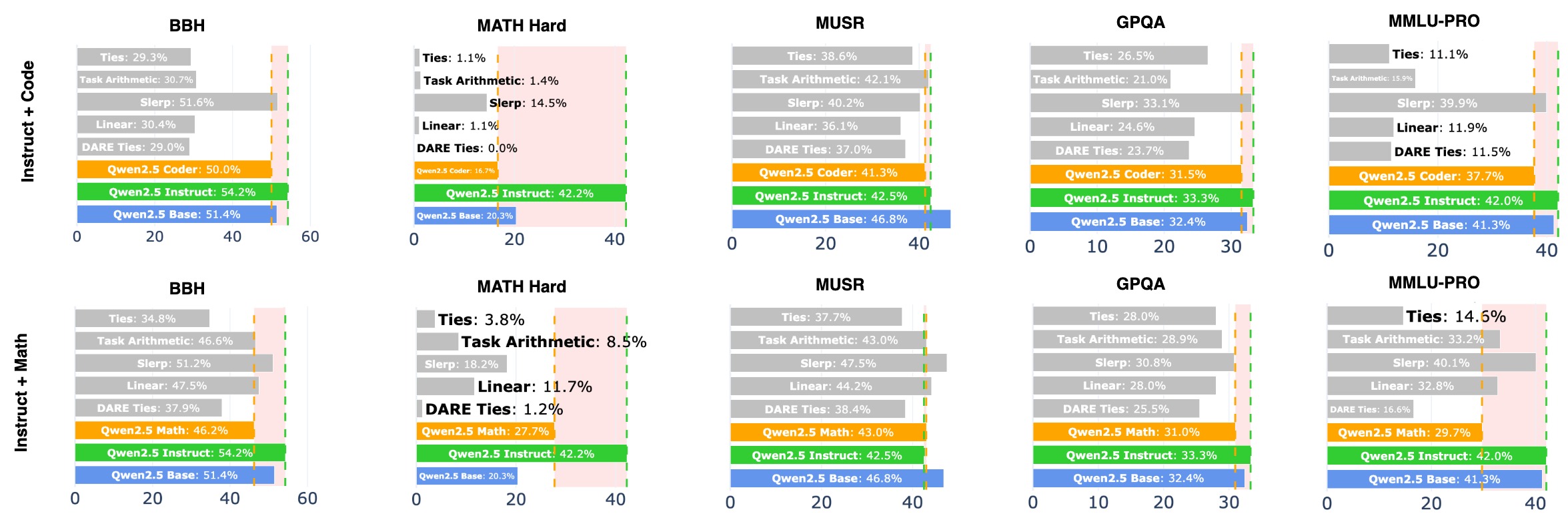}
    \caption{This horizontally grouped bar chart illustrates the absolute performance of each model across the main \texttt{Harness} tasks. The red shaded area highlights the performance gap between the instruct and math/coder models. Notably, simpler methods generally outperform more complex ones. Overall, all merged models show poorer performance compared to their parent models.}
    \label{fig:barchart_plot_harness_absolute}
\end{figure*}

\section{Evaluation Results}\label{sec:eval_results}
We merge \texttt{Instruct} with \texttt{Coder} and \texttt{Math} models from the 7B \texttt{Qwen-2.5} family \cite{qwen2025qwen25technicalreport}. 
Leveraging both the LM-Harness and Holmes frameworks reveals a significant and consistent divergence in how model merging impacts external behavior versus internal representations. While downstream tasks, the best merged models perform between the two parent models, the inherent linguistic competence encoded within the model internals tends to increase, particularly when merging \texttt{Instruct} and \texttt{Math}.

\subsection{Model Behavioral Evaluation}
We first summarize behavioral results from the LM-Harness evaluation in Figure \ref{fig:barchart_plot_harness_absolute}.

\paragraph{The behavioral compromise of model merging.}
These results consistently demonstrate that merged models can not match the performance of one of their parent models, but mostly perform between them.
As an exception, in both \texttt{Instruct + Math} and \texttt{Instruct + Coder} experiments, the merged models fail to reach at least one parent model's performance for \texttt{MATH Hard}. 
These results align with the general low performance of domain-adapted models and previous results, which underscore the importance of instruction-tuning in combination with domain adaptation \citep{Pham_OpenLLM_Operating_LLMs_2023}.

\begin{figure}
    \centering
    \includegraphics[width=1\linewidth]{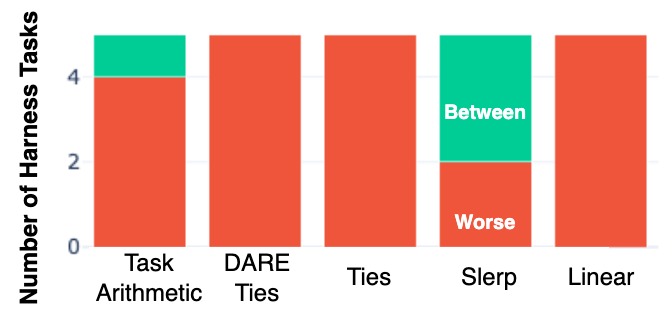}
    \caption{This stacked bar chart illustrates how each model's performance on \texttt{Leaderboard} Subtasks compares to both the \texttt{Instruct and Coder} models. Generally, most models perform either between or worse than these two baselines. However, simpler merging methods show some subtasks where they perform better."}
    \label{fig:leaderboard_harness_worser_better_between_coder}
\end{figure}

\begin{figure}
    \centering
    \includegraphics[width=1\linewidth]{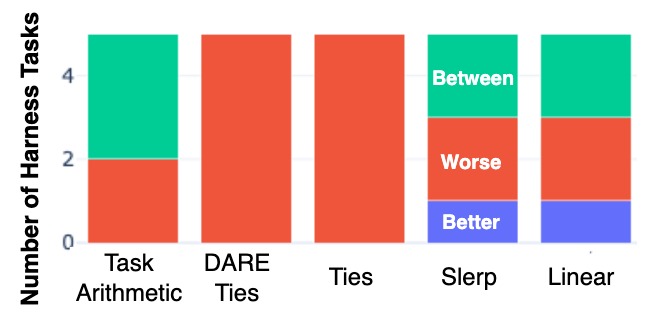}
    \caption{This stacked bar chart illustrates how each model's performance on \texttt{Leaderboard} Subtasks compares to both the \texttt{Instruct and Math} models. Generally, most models perform either between or worse than these two baselines. However, simpler merging methods show some subtasks where they perform better."}
    \label{fig:leaderboard_harness_worser_better_between_math}
\end{figure}

\paragraph{Simple model merging methods excel.}
Among the merging methods, a clear hierarchy emerges. Simpler methods consistently outperform more complex ones. As shown in Figures \ref{fig:leaderboard_harness_worser_better_between_math} and \ref{fig:leaderboard_harness_worser_better_between_coder}, \texttt{SLERP} stands out as the most effective method, frequently achieving the highest scores among the merged models and having the largest number of subtasks where its performance is "better" than both parents. \texttt{Linear} and \texttt{Task Arithmetic} follow, typically performing "between" the two parent models. In stark contrast, the more sophisticated methods, \texttt{TIES} and \texttt{DARE TIES}, which are designed to mitigate parameter interference, consistently got the poorest results. Their performance is often categorized as "worse" than both parent models, suggesting that their approach to resolving weight conflicts may be detrimental to the model's ability to perform complex, multi-step tasks.

\begin{figure*}[ht]
    \centering
    \includegraphics[width=1\linewidth]{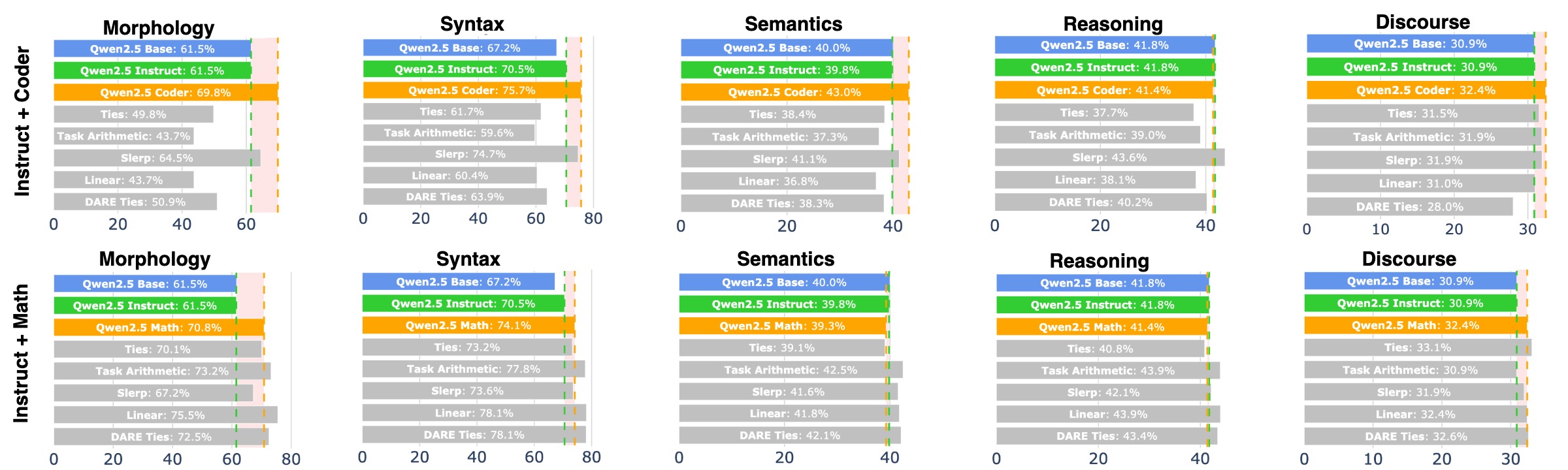}
    \caption{This horizontally grouped bar chart illustrates the absolute performance of each model across the linguistic competencies. The red-shaded area highlights the performance gap between the \texttt{Instruct and Math/Coder} models. Notably, merged models mostly outperform or perform similarly to the parent models.}
    \label{fig:barchart_plot_holmes_absolute}
\end{figure*}

\subsection{Model Internal Evaluation}
\begin{figure}
    \centering
    \includegraphics[width=1\linewidth]{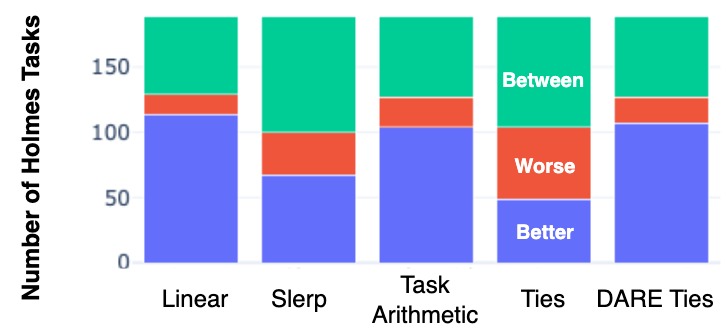}
    \caption{This stacked bar chart illustrates how each model's performance on \texttt{Holmes} tasks compares to both the \texttt{Instruct and Math} models. Generally, most models perform either better than or between these two baselines.}
    \label{fig:holmes_worser_better_between_math}
\end{figure}

\begin{figure}
    \centering
    \includegraphics[width=1\linewidth]{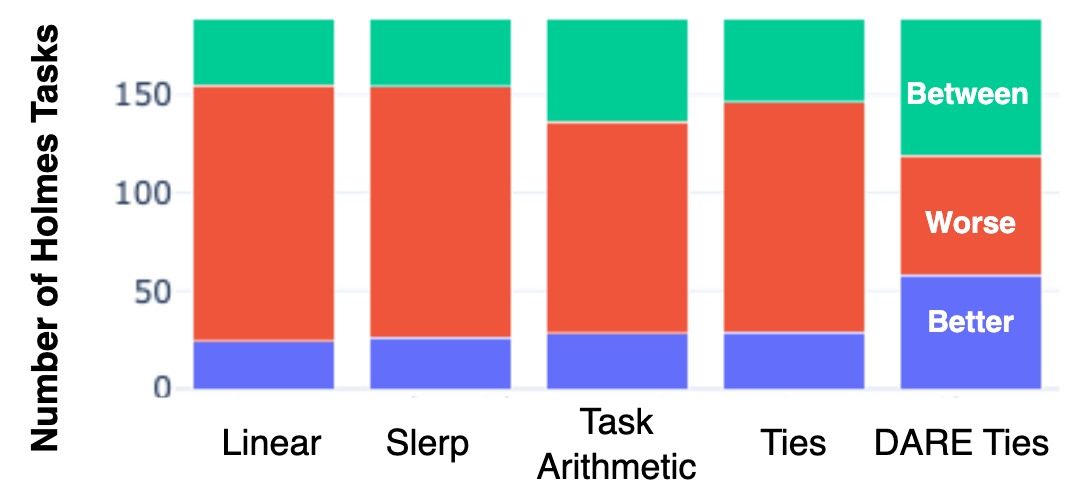}
    \caption{This stacked bar chart illustrates how each model's performance on \texttt{Holmes} tasks compares to both the \texttt{Instruct and Coder} models. Generally, most models perform either better than or between these two baselines.}
    \label{fig:holmes_worser_better_between_coder}
\end{figure}

We show in Figure \ref{fig:barchart_plot_holmes_absolute}, \ref{fig:holmes_worser_better_between_coder}, and \ref{fig:holmes_worser_better_between_math} the result of evaluating model internals regarding their inherent encoded linguistic competence. 

\paragraph{More encoded information with more training.}
More generally, we found that both instruction tuning (\texttt{Instruct}) and domain adaptation (\texttt{Coder} and \texttt{Math}) result in more information about linguistic competence compared to the \texttt{Base} model from which all originate.
This effect is particularly evident for domain adaptation, where we assume that the larger amount of data than used for instruction tuning allows LMs to capture more information about linguistic phenomena.

\paragraph{Model merging increases encoded information.}
Next, we focus on how model merging affects the model internals and find that the impact of combining models diverges from behavioral evaluations. 
Notably, we found that information about linguistic phenomena can increase when models are merged.
This effect is most pronounced for \textbf{\texttt{morphology} (word structure) and \texttt{syntax} (sentence structure)}. 
This suggests that merging can effectively combine the complementary structural knowledge from both models, resulting in internal representations with more information.

\paragraph{More information when using simpler merging methods.}
In the next step, we compare the different merging methods regarding the information encoded in the resulting models. 
Similarly, to the behavioral evaluation, we find that simpler methods (\texttt{Liner} or \texttt{SLERP}) generally preserve more information than more complex methods.
This effect is not as pronounced as when evaluating model behavior.
Notably, we find the most difference among models (parent and merged ones) in more formal phenomena, like \texttt{syntax} and \texttt{morphology}.
However, there are fewer differences in phenomena that are intuitively linked to actively using language (\texttt{reasoning}, \texttt{semantics}, or \texttt{discourse}), which we often attribute to language models.
This improvement for formal phenomena also suggests that merging can effectively transfer the information gain from seeing more tokens in further pre-training steps, as shown in \citep{waldis2024holmesbenchmarkassesslinguistic}. 

\paragraph{Adapted domain matters for merging compatibility.}
We compare the influence of the particular domain with merging into the \texttt{Instruct} model and find that information gain from model merging is not uniform. 
It is more pronounced in \texttt{Instruct + Math} than in \texttt{Instruct + Coder} experiments.
In the \texttt{Instruct + Math} experiment (Figure \ref{fig:holmes_worser_better_between_math}), nearly all merging methods produce models that outperform both the \texttt{Instruct} and \texttt{Math} models for these two phenomenon types. 
In contrast, we see that the \texttt{Coder} parent models have slightly richer model internals than the \texttt{Math} one. 
These insights suggest that model merging is sensitive to the specific domain in which it is applied. 
Specifically, we believe that the language used for \texttt{Math} adaptation is more similar to instruction tuning than that used to further pre-train the \texttt{Coder} models, resulting in better merging compatibility.

\section{Discussion}
Next, we discuss the connection results of the initial evaluation of model merging methods from a behavioral and internal perspective.

\paragraph{The efficacy of simple merging methods.}
In contrast to our expectations, we found that simple merging methods (\texttt{SLERP} and \texttt{Linear}) outperform more complex ones, such as \texttt{TIES} and \texttt{DARE TIES}, in both behavioral and internal evaluations. We hypothesize this is due to the \textbf{preservation of weight space geometry}.
This is particularly evident for \texttt{SLERP}.
We believe that finding the shortest path on the hypersphere of normalized weights respects the geometric relationships between parameters more faithfully. This geometric integrity appears to be essential for maintaining the functional coherence of the parent models.
Moreover, this supremacy underlines the need for having more holistic evaluations to assess the advantages and limitations of merging methods comprehensivly. 
We see this work, with the presented pipeline and initial results, offering the first step in this direction.

\begin{figure}
    \centering
    \includegraphics[width=1\linewidth]{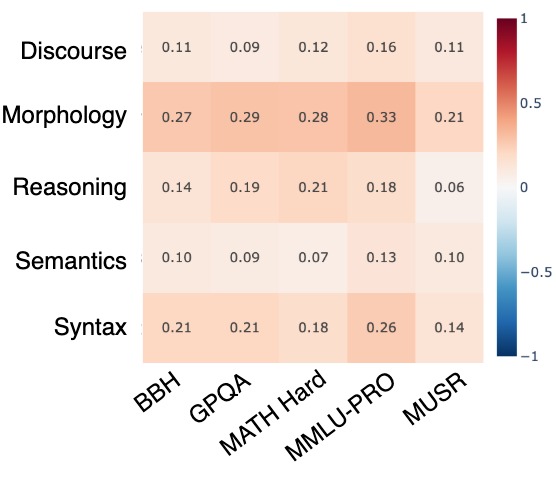}
    \caption{This plot shows a heatmap of the correlation between the linguistic competencies from \texttt{Holmes} and grouped \texttt{eval-harness} subtasks. It shows that \texttt{syntax and morphology} have the highest correlation with behavior.}
    \label{fig:correlation_between_Harness_holmes}
\end{figure}

\paragraph{Divergence between behavioral and internal evaluation of merging methods.}
The presented results indicate that behavioral and internal evaluations diverge substantially. 
While the behavioral performance of merged models decreases and is between that of the parents, the amount of encoded information within these models can increase beyond that of the parent models.
This divergence highlights how internal and behavioral evaluation offer different perspectives and underscores the need for a better understanding of how these distinct interpretability perspectives interact. 
Thus, comprehensive methods and evaluations, as presented in this work, are not only indispensable for a better understanding of various model merging methods but also essential for a more general understanding of language models, beyond those that are merged. 

\paragraph{Weak correlations of behavior and internals.}
Finally, empirically discuss results from model behavior and internals. 
We correlate the behavioral and internal results per model for tasks (eval-harness) and phenomena type (Holmes) in Figure \ref{fig:correlation_between_Harness_holmes}.
The correlations between linguistic competencies and downstream task performance are \textbf{weak to medium}. While we found the strongest correlation for \texttt{morphology} and \texttt{syntax} with \textit{eval-harness} tasks, with Pearson Correlation values ranging from 0.14 to 0.33.
These results underline, again, that \textbf{evaluating a model from a single perspective is insufficient}. A model that offers superior performance on a single leaderboard is not necessarily as internally rich as we might intuitively assume.

\section{Conclusion}

In this work, we introduce a novel evaluation pipeline that integrates model merging with model behavior and internal evaluations. 
With this novel methodology, we comprehensively assess the dynamics of different model merging methods. 
Specifically, we focused on combining instruction fine-tuned models with math and coding adapted models. 
These initial results suggest clear divergence between model behavioral and internal evaluations. 
While merged models tend to perform between the two parent models, they can encode more linguistic information than these two models, particularly for \texttt{morphology and syntax}. 
Moreover, these results also suggest that simpler merging methods often outperform more complex ones, which underlines the necessity of comprehensive evaluation to better understand whether methods offer general superiority. 

With these insights, we can directly answer our initially raised research questions: model merging affects internal representations, increasing the amount of information encoded in these representations beyond that of the parent, in a manner that differs from what behavioral evaluations suggest. 
With this divergence, we recognize that more fine-grained experiments are necessary to further study and strengthen the findings presented in this work, which are essential for gaining a deeper understanding and improving model merging methods.

\section*{Limitations}

\paragraph{Model Layers}
In this study, we focus solely on the last layer of language models to investigate what information is encoded.
While these would have expanded the scope of this work, studying how information flows through all model layers during model merging can further enhance our understanding of model merging methods, as well as of language models in general. 

\paragraph{Language Models}
This study aims to present, through the pipeline, the methodological groundwork to assess model merging methods more comprehensively. 
In this context, we present initial results to showcase the effectiveness of this pipeline and derive first insights that guide investigations of model merging. 
For this purpose, we only experimented with models of one model family (QWEN-2.5). 
However, evaluating our results alongside those of other families could strengthen our findings and also uncover further differences among models. 

\paragraph{English Language}
Given the widespread availability of evaluation resources, we limited this study to the English language. 
Thereby, the presented pipeline is not directly applicable to multilingual merging methods \citep{tao-etal-2024-unlocking}.

\section*{Acknowledgements}
We thank the anonymous reviewer for their valuable feedback and discussions. 
Andreas Waldis is supported by the Hasler Foundation Grant No. 21024.

\bibliography{anthology,custom}
\bibliographystyle{acl_natbib}

\end{document}